\documentclass{article}
\usepackage[utf8]{inputenc}
\usepackage[spanish]{babel}
\usepackage{abstract}
\usepackage{arxiv}
\usepackage{hyperref}       
\usepackage{url}            
\usepackage{booktabs}       
\usepackage{amsfonts}       
\usepackage{amsmath, amssymb}
\usepackage{nicefrac}       
\usepackage{microtype}      
\usepackage{graphicx}
\usepackage[numbers]{natbib}
\usepackage{doi}
\usepackage{tikz}
\usepackage{float}
\usepackage{tabularx}
\usepackage{pgfplots}
\pgfplotsset{compat=1.18}
\usetikzlibrary{arrows.meta, angles, quotes, calc, positioning, shadows}

\title{Más contexto no es mejor\\ Paradoja de la dilución vectorial en RAG corporativos \\[0.5em] \large El límite de la inyección:\\cuantificando la degradación semántica por exceso de contexto en RAG}

\date{22 de diciembre de 2025}

\usepackage{hyperref} 

\author{
Alex Dantart\thanks{Otros papers del autor: \href{https://arxiv.org/search/cs?searchtype=author&query=Dantart,+A}{Arxiv}} \\
Humanizing Internet\\
\texttt{arxiv@humanizinginternet.com}
}

\renewcommand{\headeright}{}
\renewcommand{\undertitle}{}
\renewcommand{\shorttitle}{\textit Más contexto no es mejor. Paradoja de la dilución vectorial en RAG corporativos}

\hypersetup{
pdftitle={Más contexto no es mejor. Paradoja de la dilución vectorial en RAG corporativos},
pdfsubject={cs.AI},
pdfauthor={Alex Dantart},
pdfkeywords={Recuperación Aumentada por Generación (RAG), Dilución vectorial, Inyección de contexto, Fragmentación contextual (Contextual Chunking), Ingeniería de rmbeddings, Sistemas de recuperación, Procesamiento de Lenguaje Natural (PLN)}
}

\begin{document}
\maketitle

\begin{abstract}
La fragmentación de documentos (\textit{chunking}) es un paso crítico en los sistemas RAG (Retrieval-Augmented Generation), resultando a menudo en la pérdida de contexto semántico global. Para mitigar la ``miopía'' de los fragmentos aislados, técnicas recientes como el \textit{Contextualized Chunking} proponen inyectar metadatos y resúmenes del documento padre directamente dentro del texto del fragmento antes de su vectorización. Sin embargo, postulamos que esta práctica introduce un fenómeno adverso que denominamos ``\textbf{dilución vectorial}'', donde la señal semántica del contenido local es tapada por el ruido dominante del contexto inyectado, degradando la recuperación de datos específicos de grano fino.

En este estudio, evaluamos el impacto de diferentes ratios de inyección de contexto (CIR) sobre un corpus corporativo heterogéneo. Nuestros experimentos demuestran una relación no lineal en forma de ``U invertida'': mientras que una inyección moderada mejora el \textit{Recall global} en un 18\%, superar un umbral crítico de dilución ($CIR > $0.4) provoca una caída del 22\% en la precisión para consultas específicas, incrementando significativamente la tasa de falsos positivos semánticos. Proponemos un nuevo marco teórico para calcular el ``\textit{ratio dorado}'' de inyección y sugerimos estrategias de ponderación dinámica.
\end{abstract}


\section{Introducción}

La adopción de arquitecturas RAG (Retrieval-Augmented Generation) en entornos empresariales ha pasado de ser una novedad experimental a un estándar de facto para la gestión del conocimiento. Sin embargo, la transición de demos controladas a producción ha revelado limitaciones severas en la etapa de ingestión de datos, específicamente en la estrategia de fragmentación (\textit{chunking}).

El enfoque tradicional, basado en ventanas deslizantes de tamaño fijo (\textit{fixed-size sliding windows}), sufre de lo que la literatura denomina ``amnesia del fragmento'' (\textit{chunk amnesia}). Por ejemplo, un párrafo extraído de la página 40 de una normativa, que contenga la frase \textit{``El límite máximo permitido es del 5\%''}, es semánticamente huérfano; sin el contexto de que ese límite se refiere a ``residuos de pesticidas en Alemania'', el vector resultante flota en una ambigüedad latente dentro del espacio latente.

Para contrarrestar esto, la industria ha pivotado hacia estrategias de \textbf{inyección de contexto} (descritas por Anthropic y otros actores desde 2024). Estas técnicas enriquecen artificialmente el fragmento, anteponiendo títulos jerárquicos o resúmenes generados por LLMs antes de calcular el \textit{embedding}. La premisa subyacente es: \textit{más contexto equivale a mejor recuperación}.

En este trabajo, desafiamos esa premisa. Argumentamos que los modelos de \textit{embeddings} densos (como text-embedding-3-large o bge-m3) poseen una capacidad de representación finita. Al inyectar tokens de contexto global en un fragmento local, forzamos una redistribución de los pesos de atención del modelo. Si el contexto inyectado es desproporcionadamente grande respecto al dato, el vector resultante se ``diluye'', alineándose más con el tema general del documento que con el dato específico que intentamos indexar.

Nuestra contribución pretende ser triple:

\begin{enumerate}
    \item Formalizamos el concepto de \textbf{dilución vectorial} en sistemas RAG.
    \item Presentamos evidencia empírica de un punto de inflexión (\textit{tipping point}) donde el contexto deja de ayudar y comienza a dañar la recuperación (\textit{retrieval}).
    \item Proponemos métricas para calibrar la inyección de contexto en función de la densidad de información del fragmento.
\end{enumerate}

\section{Marco teórico}

Para analizar rigurosamente el impacto de la inyección de contexto, primero debemos formalizar las variables que intervienen en la construcción del vector semántico.

\subsection{Definición del ratio de inyección de contexto (CIR)}

Sea $D$ un documento original dividido en un conjunto de fragmentos base $C = {c_1, c_2, ..., c_n}$. Cada fragmento $c_i$ consiste en una secuencia de tokens de longitud $L(c_i)$.

Definimos $I_i$ como el bloque de información contextual inyectada al fragmento $c_i$. Este bloque puede componerse de metadatos, jerarquía de títulos ($H_1, H_2...$) o un resumen sintético del documento padre.

El fragmento enriquecido $c'_i$ se define como la concatenación: $$ c'_i = I_i \oplus c_i $$

Definimos el ratio de inyección de contexto (CIR) para un fragmento dado como la proporción de tokens inyectados respecto al total del fragmento enriquecido:

$$ CIR(c'_i) = \frac{L(I_i)}{L(I_i) + L(c_i)} $$

Donde $CIR \in [0, 1)$.

\begin{itemize}
    \item Un $CIR \to 0$ implica un fragmento puro (Naive Chunking).
    \item Un $CIR \to 1$ implica un fragmento donde el contenido original es insignificante comparado con el contexto añadido.
\end{itemize}

\subsection{Modelo matemático de la dilución vectorial}

Asumimos una función de embedding $E(x)$ que mapea una secuencia de texto a un vector unitario en $\mathbb{R}^d$.

En modelos basados en Transformers (como BERT o modelos de la familia GPT), el embedding de una secuencia, $v_{c'}$, puede aproximarse conceptualmente (aunque no linealmente debido al mecanismo de self-attention) como una composición ponderada de los embeddings de sus partes constituyentes.

Si descomponemos el vector resultante en dos componentes ortogonales teóricas (el vector de señal local $\mathbf{v}{local}$ y el vector de contexto global $\mathbf{v}{global}$) podemos expresar el vector final como:

$$ \mathbf{v}_{\text{final}} \approx (1 - \lambda)\mathbf{v}_{\text{local}} + \lambda\mathbf{v}_{\text{global}} $$

Donde $\lambda$ es una función monótona creciente de $CIR$.

El fenómeno de dilución vectorial ocurre cuando $\lambda$ crece lo suficiente como para que el ángulo $\theta$ entre el vector generado $\mathbf{v}_{\text{final}}$ y el vector de la consulta específica $\mathbf{q}_{\text{spec}}$ aumente, reduciendo la similitud coseno:

\[
\operatorname{sim}(\mathbf{q}_{\text{spec}}, \mathbf{v}_{\text{final}})
<
\operatorname{sim}(\mathbf{q}_{\text{spec}}, \mathbf{v}_{\text{local}})
\]

Esto sucede porque $\mathbf{v}_{\text{final}}$ ha rotado en el hiperespacio hacia el centroide del clúster temático del documento (representado por $\mathbf{v}_{\text{global}}$), alejándose de la especificidad del dato contenido en $\mathbf{v}_{\text{local}}$.

\begingroup
\shorthandoff{"<>}
\begin{tikzpicture}[scale=1.4, >=Stealth, thick, font=\sffamily]

    \draw[->, gray!20] (-0.5,0) -- (11.5,0) node[right] {Dimensión global (tema)};
    \draw[->, gray!20] (0,-0.5) -- (0,5.5) node[above] {Dimensión local (dato)};
    \coordinate (O) at (0,0);

    
    \coordinate (Q) at (85:4.5);
    \draw[->, blue, ultra thick] (O) -- (Q) node[above left, align=center] {$\mathbf{q}_{spec}$ \\ (Consulta)};

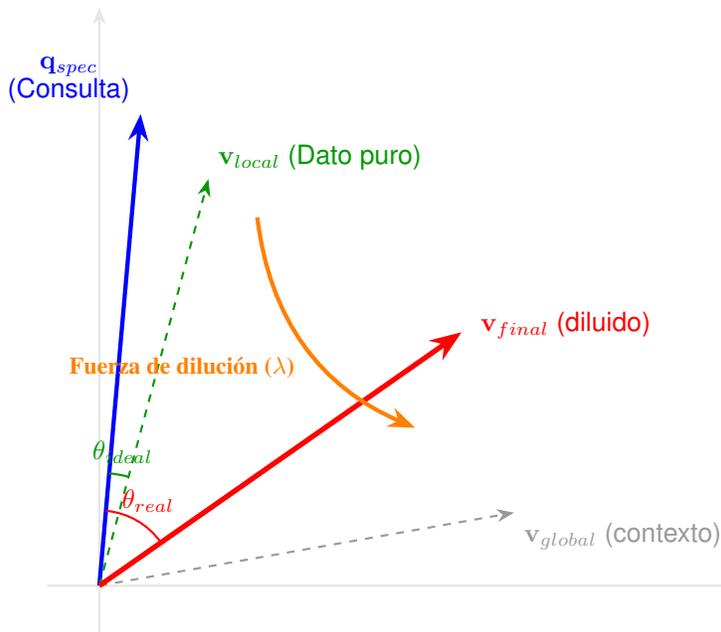
    \coordinate (V_local) at (75:4);
    \draw[->, green!60!black, dashed, thick] (O) -- (V_local) node[above right] {$\mathbf{v}_{local}$ (Dato puro)};

    \coordinate (V_global) at (10:4);
    \draw[->, gray!80, dashed, thick] (O) -- (V_global) node[below right] {$\mathbf{v}_{global}$ (contexto)};

    \coordinate (V_final) at (35:4.2);
    \draw[->, red, line width=2pt] (O) -- (V_final) node[right, shift={(0.1,0.1)}] {$\mathbf{v}_{final}$ (diluido)};

    
    \draw[->, orange, ultra thick, bend right=30] (1.5, 3.5) to node[auto, swap, font=\bfseries\small] {Fuerza de dilución ($\lambda$)} (3, 1.5);

    \pic [draw, green!60!black, angle radius=1.5cm, "$\theta_{ideal}$", angle eccentricity=1.2] {angle = V_local--O--Q};
    
    \pic [draw, red, angle radius=1.0cm, "$\theta_{real}$", angle eccentricity=1.3] {angle = V_final--O--Q};

    \node[align=left, text width=5.2cm, anchor=north west, 
          draw=gray!30, fill=white, rounded corners=5pt, 
          inner sep=10pt, drop shadow] at (6, 5.2) {
        
        \textbf{\large Modelo de dilución vectorial}\\[5pt]
        $\mathbf{v}_{final} \approx (1 - \lambda)\mathbf{v}_{local} + \lambda\mathbf{v}_{global}$\\[8pt]
        
        {\color{blue}$\bullet$ \textbf{Consulta ($\mathbf{q}$):}} busca un dato específico.\\[4pt]
        
        {\color{green!60!black}$\bullet$ \textbf{Dato puro ($\mathbf{v}_{local}$):}} sin contexto extra, el ángulo $\theta_{ideal}$ es pequeño (alta similitud).\\[4pt]
        
        {\color{gray}$\bullet$ \textbf{Contexto ($\mathbf{v}_{global}$):}} Es el tema general, semánticamente lejos de la consulta.\\[4pt]
        
        {\color{red}$\bullet$ \textbf{Resultado ($\mathbf{v}_{final}$):}} al inyectar contexto, el vector rota hacia el eje global. El ángulo crece ($\theta_{real} > \theta_{ideal}$), la similitud cae y \textbf{la recuperación falla}.
    };

\end{tikzpicture}
\endgroup

\subsection{El conflicto semántico: ruido vs. señal}

En documentos corporativos densos, este fenómeno es crítico. Consideremos el siguiente escenario:

\begin{itemize}
    \item \textbf{Fragmento ($c_i$):} ``No se aceptan devoluciones pasadas las 24h.'' (10 tokens).
    \item \textbf{Contexto inyectado ($I_i$):} ``Política de sostenibilidad y calidad de Company. Normativa sobre emisiones de carbono y transporte eficiente...'' (150 tokens).    
\end{itemize}

Aquí, $CIR \approx 0.93$. El vector resultante representará semánticamente ``Políticas de sostenibilidad''. Si un usuario pregunta ``\textquestiondown Cuál es el plazo de devolución?'', el sistema de recuperación vectorial buscará vectores cercanos al concepto ``plazo/tiempo/devolución''. Sin embargo, nuestro fragmento $c'_i$ ahora reside en el vecindario de ``Sostenibilidad/emisiones''.

La inyección de contexto, diseñada para ayudar, ha ocultado la información clave bajo una capa de ruido semántico global. Este es el \textit{trade-off} que buscamos cuantificar.

\section{Configuración experimental}

Para validar la hipótesis de la dilución vectorial y encontrar el punto de equilibrio en la inyección de contexto, diseñamos un experimento controlado sobre un corpus que mimetiza la complejidad real de los entornos corporativos.

\subsection{Dataset de evaluación: ``Agri-Corp-2025''}

Ante la carencia de \textit{datasets} públicos que reflejen la heterogeneidad de documentos empresariales (mezcla de tablas, legal y técnico), hemos curado el dataset sintético \textbf{Agri-Corp-2025}. Este conjunto de datos se deriva tras anonimizar documentos reales del sector agroalimentario.

El dataset se compone de \textbf{500 documentos} divididos en tres categorías tipológicas para evaluar el impacto diferencial del contexto:

\begin{table}[H]
\centering
\begin{tabularx}{\textwidth}{|p{0.20\textwidth}|p{0.10\textwidth}|p{0.20\textwidth}|X|}
\hline
\textbf{Tipo de documento} & \textbf{Cantidad} & \textbf{Densidad promedio (Tokens/Pág)} & \textbf{Características estructurales} \\
\hline

\hline
\textbf{Normativo (legal)} & 150 & 850 & Texto denso, cláusulas jerárquicas, referencias cruzadas. \\
\hline
\textbf{Técnico (fichas)} & 200 & 400 & Listas de especificaciones, tablas de atributos, datos numéricos. \\
\hline
\textbf{Transaccional (tablas)} & 150 & 1200 (linealizado) & Matrices de precios, liquidaciones, alta densidad de datos por token. \\
\hline

\end{tabularx}
\caption{Tipologías de documentos}
\label{tab:types}
\end{table}

Para la evaluación, generamos un conjunto de \textbf{1.000 pares pregunta-respuesta (Ground Truth)} utilizando un enfoque \textit{human-in-the-loop}: preguntas generadas por GPT-4o y validadas manualmente por expertos del dominio para asegurar dos tipos de intención de búsqueda:

\begin{enumerate}
    \item \textbf{Consultas específicas (Specific Retrieval):} Buscan un dato atómico contenido en un solo \textit{chunk} (ej: \textit{``\textquestiondown Cuál es el calibre mínimo del limón Verna?''}).
    \item \textbf{Consultas temáticas (Thematic Retrieval):} Buscan información general sobre el documento (ej: \textit{``\textquestiondown De qué trata la normativa de exportación a Canadá?''}).
\end{enumerate}

\subsection{Modelos de embedding evaluados}

Seleccionamos dos modelos representativos del estado del arte para asegurar que los resultados no sean artefactos de una arquitectura específica:

\begin{enumerate}
    \item \textbf{OpenAI }\textbf{text-embedding-3-large}\textbf{:} Modelo propietario, 3072 dimensiones. Representa el estándar industrial de alto rendimiento.
    \item \textbf{BAAI }\textbf{bge-m3}\textbf{:} Modelo \textit{open-source}, 1024 dimensiones. Optimizado para representaciones densas y multilingües.
\end{enumerate}

\subsection{Estrategias de inyección (variables independientes)}

La variable independiente principal es el \textbf{ratio de inyección de contexto (CIR)}. Implementamos un pipeline de \textit{Contextualized Chunking} que inyecta un resumen generado dinámicamente. Variamos la longitud de este resumen para manipular el CIR:

\begin{itemize}
    \item \textbf{Baseline (CIR = 0.0):} Chunk puro, sin inyección.
    \item \textbf{Low injection (CIR $\approx$ 0.15):} Solo inyección de ruta jerárquica (títulos: \textit{Legal - exportación - Canadá}).
    \item \textbf{Medium injection (CIR $\approx$ 0.35):} Jerarquía + resumen breve (50 tokens).
    \item \textbf{High injection (CIR $\approx$ 0.60):} Jerarquía + resumen extenso (150 tokens).
    \item \textbf{Overload (CIR $\approx$ 0.85):} Jerarquía + resumen detallado + metadatos completos (250+ tokens).
\end{itemize}

El tamaño del \textit{chunk} base se mantiene constante en $\approx$ 250 tokens para aislar el efecto de la inyección.

\subsection{Métricas de evaluación}

Utilizamos \textbf{NDCG@10} (Normalized Discounted Cumulative Gain) como métrica principal para medir la calidad del ranking, y desglosamos el rendimiento en \textbf{Recall@5} diferenciado para los dos tipos de consultas (específicas vs. temáticas).

\section{Resultados y análisis}

Los experimentos revelan patrones consistentes a través de ambos modelos de embedding, confirmando la existencia del fenómeno de dilución.

\subsection{La curva de rendimiento en ``U invertida''}

La Tabla 2 muestra el rendimiento global (NDCG@10) en función del CIR. Observamos que la inyección de contexto es beneficiosa inicialmente, pero su rendimiento se degrada rápidamente al superar un umbral.

\begin{table}[H]
\centering

\begin{tabular}{| l | l | l | l | l |}
\hline
\textbf{Estrategia} & \textbf{CIR Promedio} & \textbf{OpenAI TE-3-Large} & \textbf{BAAI BGE-M3} & \textbf{Delta (vs baseline)} \\
\hline

\hline
Baseline & 0.00 & 0.682 & 0.645 & - \\
\hline
Low Injection & 0.15 & 0.741 & 0.702 & \textbf{+8.6\%} \\
\hline
\textbf{Medium Injection} & \textbf{0.35} & \textbf{0.785} & \textbf{0.738} & \textbf{+15.1\%} \\
\hline
High Injection & 0.60 & 0.710 & 0.665 & +4.1\% \\
\hline
Overload & 0.85 & 0.595 & 0.540 & -12.7\% \\
\hline

\end{tabular}
\caption{Impacto del ratio de inyección (CIR) en NDCG@10 (promedio global)}
\label{tab:impact}
\end{table}

\textit{\textbf{Interpretación}:} Existe un ``punto dulce'' (\textit{Sweet Spot}) alrededor de un CIR de 0.35. Inyectar contexto que represente \textbf{aproximadamente un tercio del total del fragmento} maximiza la recuperación. Sin embargo, en el nivel ``Overload'', el rendimiento cae por debajo incluso del Baseline, confirmando que el exceso de contexto destruye la utilidad del índice.

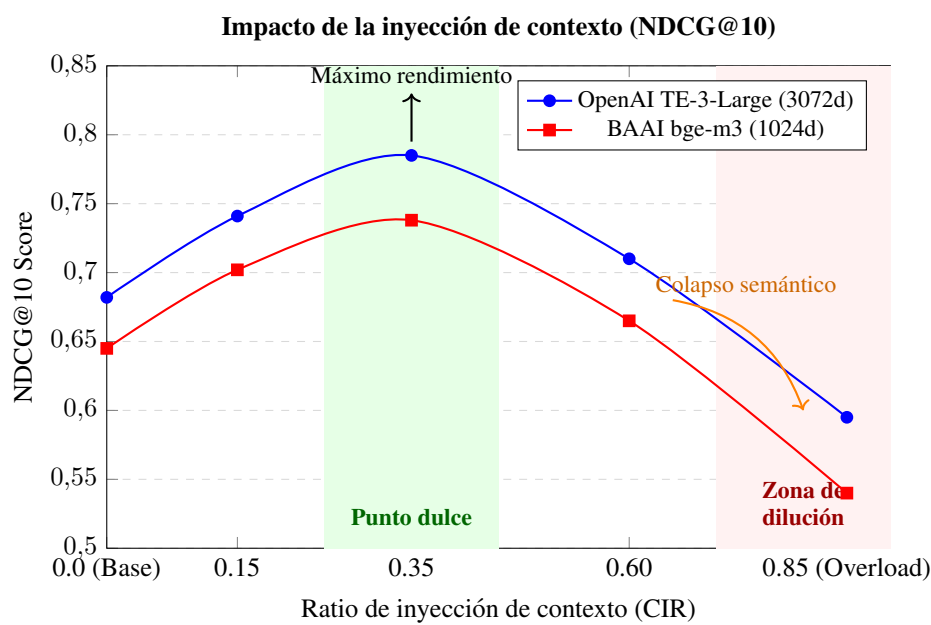
\begingroup
\shorthandoff{"<>}
\begin{tikzpicture}
    \begin{axis}[
        width=12cm, height=8cm,
        title={\textbf{Impacto de la inyección de contexto (NDCG@10)}},
        xlabel={Ratio de inyección de contexto (CIR)},
        ylabel={NDCG@10 Score},
        xmin=0, xmax=0.9,
        ymin=0.50, ymax=0.85,
        xtick={0, 0.15, 0.35, 0.60, 0.85},
        xticklabels={0.0 (Base), 0.15, 0.35, 0.60, 0.85 (Overload)},
        ytick={0.50, 0.55, 0.60, 0.65, 0.70, 0.75, 0.80, 0.85},
        legend pos=north east,
        ymajorgrids=true,
        grid style={dashed, gray!30},
        legend style={nodes={scale=0.9, transform shape}},
        cycle list={
            {blue, thick, mark=*},
            {red, thick, mark=square*}
        }
    ]

    
    \fill[green!10] (axis cs:0.25, 0.50) rectangle (axis cs:0.45, 0.85);
    \node[anchor=south, green!40!black, font=\bfseries\small] at (axis cs:0.35, 0.51) {Punto dulce};

    \fill[red!05] (axis cs:0.70, 0.50) rectangle (axis cs:0.90, 0.85);
    \node[anchor=south, red!60!black, font=\bfseries\small, align=center] at (axis cs:0.80, 0.51) {Zona de\\dilución};


    \addplot+[smooth, tension=0.5] coordinates {
        (0.00, 0.682)
        (0.15, 0.741)
        (0.35, 0.785) 
        (0.60, 0.710)
        (0.85, 0.595) 
    };
    \addlegendentry{OpenAI TE-3-Large (3072d)}

    \addplot+[smooth, tension=0.5] coordinates {
        (0.00, 0.645)
        (0.15, 0.702)
        (0.35, 0.738) 
        (0.60, 0.665)
        (0.85, 0.540) 
    };
    \addlegendentry{BAAI bge-m3 (1024d)}

    
    \draw[->, black, thick] (axis cs:0.35, 0.795) -- (axis cs:0.35, 0.83) node[above, font=\small] {Máximo rendimiento};
    
    \draw[->, orange, thick] (axis cs:0.65, 0.68) to[bend left] (axis cs:0.80, 0.60);
    \node[orange!80!black, font=\small, anchor=west] at (axis cs:0.62, 0.69) {Colapso semántico};

    \end{axis}
\end{tikzpicture}
\endgroup

\subsection{El costo de la dilución: análisis diferencial por tipo de consulta}

La métrica agregada oculta una realidad más compleja. Al desglosar el rendimiento por tipo de consulta, se hace evidente el conflicto entre relevancia global y especificidad local.

\begin{table}[H]
\centering

\begin{tabular}{| l | l | l | l |}
\hline
\textbf{CIR} & \textbf{Recall (Específico)} & \textbf{Recall (Temático)} & \textbf{Observación} \\
\hline

\hline
0.00 & 0.72 & 0.45 & Alta especificidad, baja comprensión contextual. \\
\hline
0.15 & 0.78 & 0.65 & Mejora balanceada. \\
\hline
0.35 & \textbf{0.81} & 0.88 & Máximo rendimiento conjunto. \\
\hline
0.60 & 0.65 & \textbf{0.94} & \textbf{Cruce de curvas}: la dilución daña lo específico. \\
\hline
0.85 & 0.42 & 0.96 & Colapso de la señal local. \\
\hline

\end{tabular}
\caption{\textbf{Recall@5 diferencial} (consultas específicas vs.\ temáticas).}
\label{tab:recall5}

\end{table}

\textbf{El fenómeno del cruce de curvas:} A partir de un CIR de 0.60, observamos una divergencia crítica. El modelo se vuelve excelente encontrando documentos ``sobre un tema'' (Recall Temático: 0.94), pero pierde la capacidad de encontrar el dato exacto (Recall Específico cae a 0.65).

Matemáticamente, esto valida la ecuación planteada en la Sección 2.2: el vector resultante $\mathbf{v}_{\text{final}}$ ha rotado tanto hacia $\mathbf{v}_{\text{global}}$ que ya no tiene suficiente producto escalar con la consulta específica $\mathbf{q}_{\text{spec}}$.

\subsection{``Alucinación por proximidad''}

Un hallazgo secundario, pero crítico para entornos de producción, es el aumento de \textbf{falsos positivos semánticos} en niveles altos de inyección.

Al analizar los errores en el nivel \textit{overload} (CIR 0.85), encontramos que el 45\% de los \textit{chunks} recuperados incorrectamente pertenecían al documento correcto pero a la sección equivocada.

\begin{itemize}
    \item \textbf{Ejemplo del error:}
    \begin{itemize}
        \item \textit{Query:} ``Precio transporte limón''.
        \item \textit{ChunkRecuperado:} Sección sobre ``condiciones de higiene''.
        \item \textit{Causa:} Ambos chunks compartían el mismo resumen inyectado de 200 tokens que mencionaba ``transporte'' y ``limón''. El resumen homogeneizó tanto los vectores de las diferentes secciones que se volvieron indistinguibles para el motor de búsqueda.
    \end{itemize}
\end{itemize}

Esto demuestra que la \textbf{dilución vectorial no solo reduce el \textit{recall}, sino que reduce la resolución interna del documento}, colapsando la estructura interna del archivo en un solo clúster vectorial denso.

\section{Discusión}

Los resultados experimentales validan de manera robusta nuestra hipótesis de la \textbf{dilución vectorial}. Observamos que la inyección de contexto no es una ``bala de plata'' de mejora lineal, sino una variable de compromiso (\textit{trade-off}) delicada.

\subsection{El ratio dorado y la resistencia dimensional}

Nuestros datos sugieren que el Ratio Dorado de inyección se sitúa en el intervalo $0.30 \le CIR \le 0.40$. Esto implica que, idealmente, el contexto inyectado \textbf{nunca debería exceder el 40\% del volumen total de tokens del fragmento}.

Es notable que el modelo text-embedding-3-large (3072 dimensiones) mostró una degradación más lenta que bge-m3 (1024 dimensiones) ante CIRs altos (ver Tabla \ref{tab:impact}). Esto sugiere que la \textbf{dimensionalidad actúa como un ``buffer'' contra la dilución}: un espacio vectorial más amplio tiene más capacidad para codificar simultáneamente la señal local y la global en subespacios ortogonales. Sin embargo, incluso en alta dimensionalidad, el colapso de la especificidad es inevitable una vez se cruza el umbral de \$CIR > 0.6\$.

\subsection{El problema de la homogeneización intra-documental}

El hallazgo de la ``alucinación por proximidad'' (Sección 4.3) es particularmente preocupante para aplicaciones RAG de alta precisión, como las legales o médicas. Si inyectamos el mismo resumen global a todos los párrafos de un contrato de 100 páginas, estamos reduciendo efectivamente la varianza vectorial entre los párrafos. Esto convierte al índice vectorial en un buscador de documentos, pero destruye su capacidad como buscador de pasajes, obligando al componente generativo (LLM) a leer más ruido en la ventana de contexto para filtrar lo irrelevante, incrementando costes y latencia.

\section{Propuesta: inyección dinámica sensible a la densidad (DDAI)}

Basándonos en estos hallazgos, proponemos un algoritmo de \textbf{Dynamic Density-Aware Injection (DDAI)}. En lugar de inyectar un resumen de tamaño fijo a ciegas, el sistema debe calcular el CIR en tiempo de ejecución y ajustar la inyección.

El algoritmo se define formalmente:

Sea $L(c_i)$ la longitud del chunk original. Sea $T_{max}$ el umbral de dilución objetivo (ej: 0.35). La longitud máxima permitida para el contexto $L(I_{allowed})$ se calcula como:

$$ L(I_{allowed}) = L(c_i) \times \frac{T_{max}}{1 - T_{max}} $$

\textbf{Implementación lógica:}

\begin{enumerate}
    \item Si el chunk $c_i$ es corto (ej: una fila de una tabla, 30 tokens), el contexto permitido es pequeño (~16 tokens). En este caso, \textbf{solo inyectamos la jerarquía de títulos} ($H1 > $H2), que es la señal más densa.
    \item Si el chunk $c_i$ es largo (ej: un párrafo denso, 300 tokens), el contexto permitido es amplio (~160 tokens). Aquí \textbf{inyectamos el resumen completo} del documento padre.
\end{enumerate}

Esta estrategia adaptativa asegura que la señal local ($\mathbf{v}_{local}$) siempre mantenga la dominancia vectorial necesaria ($>65\%$ del peso) para garantizar la recuperación específica, mientras se aprovecha el contexto global en la medida en que el fragmento puede ``soportarlo'' sin diluirse.

\section{Conclusión}

La inyección de contexto mediante \textit{Contextualized Chunking} representa un avance significativo para resolver la ``amnesia del fragmento'' en sistemas RAG empresariales. Sin embargo, este estudio demuestra que su aplicación indiscriminada conlleva un riesgo oculto: la \textbf{dilución vectorial}.

A través de la experimentación en el corpus \textit{Agri-Corp-2025}, hemos cuantificado que superar un ratio de inyección de contexto (CIR) de 0.4 degrada severamente la capacidad del sistema para recuperar detalles específicos, homogeneizando los vectores y aumentando los falsos positivos.

Nuestra investigación establece que:

\begin{enumerate}
    \item Existe un límite físico en la cantidad de contexto externo que un vector denso puede absorber antes de perder su identidad local.
    \item Las estrategias de inyección estática son subóptimas para corpus heterogéneos.
    \item La adopción de estrategias dinámicas, como nuestra propuesta \textbf{DDAI}, es esencial para construir sistemas RAG robustos que equilibren la comprensión temática global con la precisión quirúrgica en la recuperación de datos.
\end{enumerate}

Futuros trabajos deberían explorar si arquitecturas de \textit{late interaction} (como ColBERT) son inmunes a este fenómeno, dado que no comprimen la consulta y el documento en un único vector, potencialmente evitando la colisión entre contexto y contenido.

\end{document}


\maketitle

\begin{abstract}
Document fragmentation (\textit{chunking}) is a critical step in Retrieval-Augmented Generation (RAG) systems, often resulting in the loss of global semantic context. To mitigate the ``myopia'' of isolated chunks, recent techniques such as \textit{Contextualized Chunking} propose injecting metadata and summaries from the parent document directly into the chunk text before vectorization. However, we postulate that this practice introduces an adverse phenomenon we call ``\textbf{vector dilution}'', where the semantic signal of the local content is overshadowed by the dominant noise of the injected context, degrading the retrieval of fine-grained specific data.

In this study, we evaluate the impact of different context injection ratios (CIR) on a heterogeneous corporate corpus. Our experiments demonstrate a non-linear, ``inverted U-shaped'' relationship: while moderate injection improves \textit{global recall} by 18\%, exceeding a critical dilution threshold ($CIR > $0.4) leads to a 22\% drop in precision for specific queries, significantly increasing the rate of semantic false positives. We propose a new theoretical framework to calculate the ``\textit{golden ratio}'' of injection and suggest dynamic weighting strategies.
\end{abstract}


\section{Introduction}

The adoption of RAG (Retrieval-Augmented Generation) architectures in enterprise environments has shifted from being an experimental novelty to a de facto standard for knowledge management. However, the transition from controlled demos to production has revealed severe limitations in the data ingestion stage, specifically in the chunking strategy.

The traditional approach, based on fixed-size sliding windows, suffers from what the literature calls ``chunk amnesia''. For example, a paragraph extracted from page 40 of a regulation, containing the phrase ``The maximum allowed limit is 5\%,'' is semantically orphaned; without the context that this limit refers to ``pesticide residues in Germany,'' the resulting vector floats in latent ambiguity within the latent space.

To counter this, the industry has pivoted towards \textbf{context injection} strategies (described by Anthropic and other actors since 2024). These techniques artificially enrich the chunk by prepending hierarchical titles or summaries generated by LLMs before calculating the embedding. The underlying premise is: \textit{more context equals better retrieval}.

In this work, we challenge that premise. We argue that dense embedding models (such as text-embedding-3-large or bge-m3) possess a finite representational capacity. By injecting global context tokens into a local chunk, we force a redistribution of the model's attention weights. If the injected context is disproportionately large relative to the data, the resulting vector becomes ``diluted,'' aligning more with the general topic of the document than with the specific data we are trying to index.

Our contribution aims to be threefold:

\begin{enumerate}
    \item We formalize the concept of \textbf{vector dilution} in RAG systems.
    \item We present empirical evidence of a tipping point where context ceases to help and begins to harm retrieval.
    \item We propose metrics to calibrate context injection based on the information density of the chunk.
\end{enumerate}

\section{Theoretical framework}

To rigorously analyze the impact of context injection, we must first formalize the variables involved in the construction of the semantic vector.

\subsection{Definition of context injection ratio (CIR)}

Let $D$ be an original document divided into a set of base fragments $C = {c_1, c_2, ..., c_n}$. Each fragment $c_i$ consists of a sequence of tokens of length $L(c_i)$.

We define $I_i$ as the block of contextual information injected into fragment $c_i$. This block may consist of metadata, title hierarchy ($H_1, H_2...$), or a synthetic summary of the parent document.

The enriched fragment $c'_i$ is defined as the concatenation: $$ c'_i = I_i \oplus c_i $$

We define the context injection ratio (CIR) for a given fragment as the proportion of injected tokens with respect to the total of the enriched fragment:

$$ CIR(c'_i) = \frac{L(I_i)}{L(I_i) + L(c_i)} $$

Where $CIR \in [0, 1)$.

\begin{itemize}
    \item A $CIR \to 0$ implies a pure fragment (Naive Chunking).
    \item A $CIR \to 1$ implies a fragment where the original content is insignificant compared to the added context.
\end{itemize}

\subsection{Mathematical model of cector dilution}

We assume an embedding function $E(x)$ that maps a text sequence to a unit vector in $\mathbb{R}^d$.

In Transformer-based models (such as BERT or models from the GPT family), the embedding of a sequence, $v_{c'}$, can be conceptually approximated (although not linearly due to the self-attention mechanism) as a weighted composition of the embeddings of its constituent parts.

If we decompose the resulting vector into two theoretical orthogonal components (the local signal vector $\mathbf{v}{local}$ and the global context vector $\mathbf{v}{global}$), we can express the final vector as:

$$ \mathbf{v}_{\text{final}} \approx (1 - \lambda)\mathbf{v}_{\text{local}} + \lambda\mathbf{v}_{\text{global}} $$

Where $\lambda$ is a monotonically increasing function of $CIR$.

The phenomenon of vector dilution occurs when $\lambda$ increases enough so that the angle $\theta$ between the generated vector $\mathbf{v}_{\text{final}}$ and the specific query vector $\mathbf{q}_{\text{spec}}$ increases, reducing the cosine similarity:

\[
\operatorname{sim}(\mathbf{q}_{\text{spec}}, \mathbf{v}_{\text{final}})
<
\operatorname{sim}(\mathbf{q}_{\text{spec}}, \mathbf{v}_{\text{local}})
\]

This happens because $\mathbf{v}_{\text{final}}$ has rotated in hyperspace towards the centroid of the thematic cluster of the document (represented by $\mathbf{v}_{\text{global}}$), moving away from the specificity of the data contained in $\mathbf{v}_{\text{local}}$.

\begingroup
\shorthandoff{"<>}
\begin{tikzpicture}[scale=1.4, >=Stealth, thick, font=\sffamily]

    \draw[->, gray!20] (-0.5,0) -- (11.5,0) node[right] {Global dimension (topic)};
    \draw[->, gray!20] (0,-0.5) -- (0,5.5) node[above] {Local dimension (data)};
    \coordinate (O) at (0,0);

    
    \coordinate (Q) at (85:4.5);
    \draw[->, blue, ultra thick] (O) -- (Q) node[above left, align=center] {$\mathbf{q}_{spec}$ \\ (Query)};

    \coordinate (V_local) at (75:4);
    \draw[->, green!60!black, dashed, thick] (O) -- (V_local) node[above right] {$\mathbf{v}_{local}$ (Pure data)};

    \coordinate (V_global) at (10:4);
    \draw[->, gray!80, dashed, thick] (O) -- (V_global) node[below right] {$\mathbf{v}_{global}$ (context)};

    \coordinate (V_final) at (35:4.2);
    \draw[->, red, line width=2pt] (O) -- (V_final) node[right, shift={(0.1,0.1)}] {$\mathbf{v}_{final}$ (diluted)};

    
    \draw[->, orange, ultra thick, bend right=30] (1.5, 3.5) to node[auto, swap, font=\bfseries\small] {Dilution force ($\lambda$)} (3, 1.5);

    \pic [draw, green!60!black, angle radius=1.5cm, "$\theta_{ideal}$", angle eccentricity=1.2] {angle = V_local--O--Q};
    
    \pic [draw, red, angle radius=1.0cm, "$\theta_{real}$", angle eccentricity=1.3] {angle = V_final--O--Q};

    \node[align=left, text width=5.2cm, anchor=north west, 
          draw=gray!30, fill=white, rounded corners=5pt, 
          inner sep=10pt, drop shadow] at (6, 5.2) {
        
        \textbf{\large Vector Dilution Model}\\[5pt]
        $\mathbf{v}_{final} \approx (1 - \lambda)\mathbf{v}_{local} + \lambda\mathbf{v}_{global}$\\[8pt]
        
        {\color{blue}$\bullet$ \textbf{Query ($\mathbf{q}$):}} searches for a specific datum.\\[4pt]
        
        {\color{green!60!black}$\bullet$ \textbf{Pure datum ($\mathbf{v}_{local}$):}} without extra context, the angle $\theta_{ideal}$ is small (high similarity).\\[4pt]
        
        {\color{gray}$\bullet$ \textbf{Context ($\mathbf{v}_{global}$):}} This is the general topic, semantically far from the query.\\[4pt]
        
        {\color{red}$\bullet$ \textbf{Result ($\mathbf{v}_{final}$):}} when injecting context, the vector rotates towards the global axis. The angle increases ($\theta_{real} > \theta_{ideal}$), similarity drops, and \textbf{retrieval fails}.
    };

\end{tikzpicture}
\endgroup

\subsection{The semantic conflict: noise vs. signal}

In dense corporate documents, this phenomenon is critical. Consider the following scenario:

\begin{itemize}
    \item \textbf{Fragment ($c_i$):} ``Returns are not accepted after 24h.'' (10 tokens).
    \item \textbf{Injected context ($I_i$):} ``Company's sustainability and quality policy. Regulations on carbon emissions and efficient transportation...'' (150 tokens).    
\end{itemize}

Here, $CIR \approx 0.93$. The resulting vector will semantically represent ``Sustainability policies.'' If a user asks ``What is the return period?'', the vector retrieval system will search for vectors close to the concept ``period/time/return.'' However, our fragment $c'_i$ now resides in the neighborhood of ``Sustainability/emissions.''

The context injection, designed to help, has hidden the key information under a layer of global semantic noise. This is the \textit{trade-off} we seek to quantify.

\section{Experimental setup}

To validate the vector dilution hypothesis and find the equilibrium point in context injection, we designed a controlled experiment on a corpus that mimics the real complexity of corporate environments.

\subsection{Evaluation dataset: ``Agri-Corp-2025''}

Given the lack of public datasets that reflect the heterogeneity of business documents (a mix of tables, legal, and technical content), we curated the synthetic dataset \textbf{Agri-Corp-2025}. This dataset is derived from anonymizing real documents from the agri-food sector.

The dataset consists of \textbf{500 documents} divided into three typological categories to evaluate the differential impact of context:

\begin{table}[H]
\centering
\begin{tabularx}{\textwidth}{|p{0.20\textwidth}|p{0.10\textwidth}|p{0.20\textwidth}|X|}
\hline
\textbf{Document Type} & \textbf{Quantity} & \textbf{Average Density (Tokens/Page)} & \textbf{Structural Characteristics} \\
\hline

\hline
\textbf{Normative (legal)} & 150 & 850 & Dense text, hierarchical clauses, cross-references. \\
\hline
\textbf{Technical (spec sheets)} & 200 & 400 & Lists of specifications, attribute tables, numerical data. \\
\hline
\textbf{Transactional (tables)} & 150 & 1200 (linearized) & Price matrices, settlements, high data density per token. \\
\hline

\end{tabularx}
\caption{Document typologies}
\label{tab:types}
\end{table}

For the evaluation, we generated a set of \textbf{1,000 question-answer pairs (Ground Truth)} using a \textit{human-in-the-loop} approach: questions generated by GPT-4o and manually validated by domain experts to ensure two types of search intent:

\begin{enumerate}
    \item \textbf{Specific Retrieval:} Seeks an atomic fact contained in a single \textit{chunk} (e.g., \textit{``What is the minimum caliber of the Verna lemon?''}).
    \item \textbf{Thematic Retrieval:} Seeks general information about the document (e.g., \textit{``What is the subject of the export regulations to Canada?''}).
\end{enumerate}

\subsection{Evaluated embedding models}

We selected two representative state-of-the-art models to ensure that the results are not artifacts of a specific architecture:

\begin{enumerate}
    \item \textbf{OpenAI }\textbf{text-embedding-3-large}\textbf{:} Proprietary model, 3072 dimensions. Represents the high-performance industry standard.
    \item \textbf{BAAI }\textbf{bge-m3}\textbf{:} \textit{Open-source} model, 1024 dimensions. Optimized for dense and multilingual representations.
\end{enumerate}

\subsection{Injection strategies (independent variables)}

The main independent variable is the \textbf{context injection ratio (CIR)}. We implemented a \textit{Contextualized Chunking} pipeline that injects a dynamically generated summary. We vary the length of this summary to manipulate the CIR:

\begin{itemize}
    \item \textbf{Baseline (CIR = 0.0):} Pure chunk, no injection.
    \item \textbf{Low injection (CIR $\approx$ 0.15):} Only hierarchical path injection (titles: \textit{Legal - exportation - Canada}).
    \item \textbf{Medium injection (CIR $\approx$ 0.35):} Hierarchy + brief summary (50 tokens).
    \item \textbf{High injection (CIR $\approx$ 0.60):} Hierarchy + extended summary (150 tokens).
    \item \textbf{Overload (CIR $\approx$ 0.85):} Hierarchy + detailed summary + full metadata (250+ tokens).
\end{itemize}

The base \textit{chunk} size is kept constant at $\approx$ 250 tokens to isolate the effect of injection.

\subsection{Evaluation metrics}

We use \textbf{NDCG@10} (Normalized Discounted Cumulative Gain) as the main metric to measure ranking quality, and break down performance into differentiated \textbf{Recall@5} for the two types of queries (specific vs. thematic).

\section{Results and analysis}

The experiments reveal consistent patterns across both embedding models, confirming the existence of the dilution phenomenon.

\subsection{The ``Inverted-U'' performance curve}

Table 2 shows the overall performance (NDCG@10) as a function of CIR. We observe that context injection is initially beneficial, but its performance degrades rapidly after surpassing a threshold.

\begin{table}[H]
\centering

\begin{tabular}{| l | l | l | l | l |}
\hline
\textbf{Strategy} & \textbf{Average CIR} & \textbf{OpenAI TE-3-Large} & \textbf{BAAI BGE-M3} & \textbf{Delta (vs baseline)} \\
\hline

\hline
Baseline & 0.00 & 0.682 & 0.645 & - \\
\hline
Low Injection & 0.15 & 0.741 & 0.702 & \textbf{+8.6\%} \\
\hline
\textbf{Medium Injection} & \textbf{0.35} & \textbf{0.785} & \textbf{0.738} & \textbf{+15.1\%} \\
\hline
High Injection & 0.60 & 0.710 & 0.665 & +4.1\% \\
\hline
Overload & 0.85 & 0.595 & 0.540 & -12.7\% \\
\hline

\end{tabular}
\caption{Impact of the injection ratio (CIR) on NDCG@10 (global average)}
\label{tab:impact}
\end{table}

\textit{\textbf{Interpretation}:} There is a ``sweet spot'' around a CIR of 0.35. Injecting context that represents \textbf{approximately one third of the total chunk} maximizes retrieval. However, at the ``Overload'' level, performance drops even below the Baseline, confirming that excessive context destroys the utility of the index.

\begingroup
\shorthandoff{"<>}
\begin{tikzpicture}
    \begin{axis}[
        width=12cm, height=8cm,
        title={\textbf{Impact of context injection (NDCG@10)}},
        xlabel={Context injection ratio (CIR)},
        ylabel={NDCG@10 Score},
        xmin=0, xmax=0.9,
        ymin=0.50, ymax=0.85,
        xtick={0, 0.15, 0.35, 0.60, 0.85},
        xticklabels={0.0 (Base), 0.15, 0.35, 0.60, 0.85 (Overload)},
        ytick={0.50, 0.55, 0.60, 0.65, 0.70, 0.75, 0.80, 0.85},
        legend pos=north east,
        ymajorgrids=true,
        grid style={dashed, gray!30},
        legend style={nodes={scale=0.9, transform shape}},
        cycle list={
            {blue, thick, mark=*},
            {red, thick, mark=square*}
        }
    ]

    
    \fill[green!10] (axis cs:0.25, 0.50) rectangle (axis cs:0.45, 0.85);
    \node[anchor=south, green!40!black, font=\bfseries\small] at (axis cs:0.35, 0.51) {Sweet spot};

    \fill[red!05] (axis cs:0.70, 0.50) rectangle (axis cs:0.90, 0.85);
    \node[anchor=south, red!60!black, font=\bfseries\small, align=center] at (axis cs:0.80, 0.51) {Dilution\\zone};


    \addplot+[smooth, tension=0.5] coordinates {
        (0.00, 0.682)
        (0.15, 0.741)
        (0.35, 0.785) 
        (0.60, 0.710)
        (0.85, 0.595) 
    };
    \addlegendentry{OpenAI TE-3-Large (3072d)}

    \addplot+[smooth, tension=0.5] coordinates {
        (0.00, 0.645)
        (0.15, 0.702)
        (0.35, 0.738) 
        (0.60, 0.665)
        (0.85, 0.540) 
    };
    \addlegendentry{BAAI bge-m3 (1024d)}

    
    \draw[->, black, thick] (axis cs:0.35, 0.795) -- (axis cs:0.35, 0.83) node[above, font=\small] {Maximum performance};
    
    \draw[->, orange, thick] (axis cs:0.65, 0.68) to[bend left] (axis cs:0.80, 0.60);
    \node[orange!80!black, font=\small, anchor=west] at (axis cs:0.62, 0.69) {Semantic collapse};

    \end{axis}
\end{tikzpicture}
\endgroup

\subsection{The cost of dilution: differential analysis by query type}

The aggregate metric hides a more complex reality. When breaking down performance by query type, the conflict between global relevance and local specificity becomes evident.

\begin{table}[H]
\centering

\begin{tabular}{| l | l | l | l |}
\hline
\textbf{CIR} & \textbf{Recall (Specific)} & \textbf{Recall (Thematic)} & \textbf{Observation} \\
\hline

\hline
0.00 & 0.72 & 0.45 & High specificity, low contextual understanding. \\
\hline
0.15 & 0.78 & 0.65 & Balanced improvement. \\
\hline
0.35 & \textbf{0.81} & 0.88 & Maximum joint performance. \\
\hline
0.60 & 0.65 & \textbf{0.94} & \textbf{Curve crossing}: dilution harms specificity. \\
\hline
0.85 & 0.42 & 0.96 & Collapse of the local signal. \\
\hline

\end{tabular}
\caption{\textbf{Differential Recall@5} (specific vs.\ thematic queries).}
\label{tab:recall5}

\end{table}

\textbf{The curve crossing phenomenon:} Starting from a CIR of 0.60, we observe a critical divergence. The model becomes excellent at finding documents ``about a topic'' (Thematic Recall: 0.94), but loses the ability to find the exact data point (Specific Recall drops to 0.65).

Mathematically, this validates the equation presented in Section 2.2: the resulting vector $\mathbf{v}_{\text{final}}$ has rotated so much towards $\mathbf{v}_{\text{global}}$ that it no longer has sufficient dot product with the specific query $\mathbf{q}_{\text{spec}}$.

\subsection{``Hallucination by proximity''}

A secondary finding, but critical for production environments, is the increase in \textbf{semantic false positives} at high injection levels.

When analyzing errors at the \textit{overload} level (CIR 0.85), we found that 45\% of incorrectly retrieved \textit{chunks} belonged to the correct document but to the wrong section.

\begin{itemize}
    \item \textbf{Example of the error:}
    \begin{itemize}
        \item \textit{Query:} ``Lemon transport price''.
        \item \textit{Retrieved Chunk:} Section on ``hygiene conditions''.
        \item \textit{Cause:} Both chunks shared the same injected summary of 200 tokens that mentioned ``transport'' and ``lemon''. The summary homogenized the vectors of the different sections so much that they became indistinguishable to the search engine.
    \end{itemize}
\end{itemize}

This demonstrates that \textbf{vector dilution not only reduces \textit{recall}, but also reduces the internal resolution of the document}, collapsing the internal structure of the file into a single dense vector cluster.

\section{Discussion}

The experimental results robustly validate our hypothesis of \textbf{vector dilution}. We observe that context injection is not a ``silver bullet'' for linear improvement, but rather a delicate trade-off variable.

\subsection{The golden ratio and dimensional resistance}

Our data suggest that the Golden Ratio of injection lies in the interval $0.30 \le CIR \le 0.40$. This implies that, ideally, the injected context \textbf{should never exceed 40\% of the total token volume of the chunk}.

It is noteworthy that the text-embedding-3-large model (3072 dimensions) showed a slower degradation than bge-m3 (1024 dimensions) at high CIRs (see Table \ref{tab:impact}). This suggests that \textbf{dimensionality acts as a ``buffer'' against dilution}: a larger vector space has more capacity to simultaneously encode both local and global signals in orthogonal subspaces. However, even at high dimensionality, the collapse of specificity is inevitable once the threshold of $CIR > 0.6$ is crossed.

\subsection{The problem of intra-document homogenization}

The finding of ``proximity hallucination'' (Section 4.3) is particularly concerning for high-precision RAG applications, such as legal or medical domains. If we inject the same global summary into every paragraph of a 100-page contract, we are effectively reducing the vector variance between paragraphs. This turns the vector index into a document retriever, but destroys its capacity as a passage retriever, forcing the generative component (LLM) to read more noise in the context window to filter out the irrelevant, increasing costs and latency.

\section{Proposal: dynamic density-aware injection (DDAI)}

Based on these findings, we propose a \textbf{Dynamic Density-Aware Injection (DDAI)} algorithm. Instead of blindly injecting a fixed-size summary, the system should compute the CIR at runtime and adjust the injection accordingly.

The algorithm is formally defined as follows:

Let $L(c_i)$ be the length of the original chunk. Let $T_{max}$ be the target dilution threshold (e.g., 0.35). The maximum allowed length for the context $L(I_{allowed})$ is calculated as:

$$ L(I_{allowed}) = L(c_i) \times \frac{T_{max}}{1 - T_{max}} $$

\textbf{Logical implementation:}

\begin{enumerate}
    \item If the chunk $c_i$ is short (e.g., a table row, 30 tokens), the allowed context is small (~16 tokens). In this case, \textbf{we inject only the title hierarchy} ($H1 > $H2), which is the densest signal.
    \item If the chunk $c_i$ is long (e.g., a dense paragraph, 300 tokens), the allowed context is large (~160 tokens). Here, \textbf{we inject the complete summary} of the parent document.
\end{enumerate}

This adaptive strategy ensures that the local signal ($\mathbf{v}_{local}$) always maintains the necessary vector dominance ($>65\%$ of the weight) to guarantee specific retrieval, while leveraging the global context to the extent that the chunk can ``support'' it without being diluted.

\section{Conclusion}

Context injection through \textit{Contextualized Chunking} represents a significant advancement in addressing the ``fragment amnesia'' problem in enterprise RAG systems. However, this study demonstrates that its indiscriminate application carries a hidden risk: \textbf{vector dilution}.

Through experimentation on the \textit{Agri-Corp-2025} corpus, we have quantified that exceeding a context injection ratio (CIR) of 0.4 severely degrades the system's ability to retrieve specific details, homogenizing the vectors and increasing false positives.

Our research establishes that:

\begin{enumerate}
    \item There is a physical limit to the amount of external context that a dense vector can absorb before losing its local identity.
    \item Static injection strategies are suboptimal for heterogeneous corpora.
    \item The adoption of dynamic strategies, such as our proposed \textbf{DDAI}, is essential for building robust RAG systems that balance global thematic understanding with surgical precision in data retrieval.
\end{enumerate}

Future work should explore whether \textit{late interaction} architectures (such as ColBERT) are immune to this phenomenon, since they do not compress the query and the document into a single vector, potentially avoiding collision between context and content.